\def\year{2022}\relax
\documentclass[letterpaper]{article} 
\usepackage{aaai22}  
\usepackage{times}  
\usepackage{helvet}  
\usepackage{courier}  
\usepackage[hyphens]{url}  
\usepackage{graphicx} 
\usepackage{natbib}  
\usepackage{caption} 
\DeclareCaptionStyle{ruled}{labelfont=normalfont,labelsep=colon,strut=off} 
\usepackage{algorithm}

\usepackage{newfloat}
\usepackage{listings}
\floatstyle{ruled}
\newfloat{listing}{tb}{lst}{}
\floatname{listing}{Listing}
\usepackage{svg}
\usepackage{subcaption}
\usepackage{amsmath}
\usepackage{amsfonts}
\usepackage{nicefrac}
\usepackage{amsthm}
\usepackage[noend]{algpseudocode}

\newtheorem{lemma}{Lemma}
\newtheorem{definition}{Definition}
\newtheorem{theorem}{Theorem}
\newtheorem{corollary}{Corollary}

\title{Unbiased Single-scale and Multi-scale Quantizers for Distributed Optimization}
\author {
    Vineeth S 
}
\affiliations{
    Indian Institute of Science\\
    Bangalore, India\\
    vineeths@iisc.ac.in
}

\begin{document}

\maketitle

\begin{abstract}
Massive amounts of data have made the training of large-scale machine learning models on a single worker inefficient. Distributed machine learning methods such as Parallel-SGD and federated learning have received significant interest as a solution to tackle this problem. However, the performance of distributed learning systems does not scale linearly with the number of workers due to the high network communication cost for synchronizing gradients and parameters. Researchers have proposed techniques such as quantization and sparsification to alleviate this problem by compressing the gradients. Most of the compression schemes result in compressed gradients that cannot be directly aggregated with efficient protocols such as all-reduce.	In this paper, we present a set of all-reduce compatible gradient compression schemes which significantly reduce the communication overhead while maintaining the performance of vanilla SGD. Our compression methods perform better than the in-built methods currently offered by the deep learning frameworks. Code will be available at the repository: https://github.com/vineeths96/Gradient-Compression.
\end{abstract}

\noindent Deep learning models trained on large amounts of data have shown superhuman performance in tasks such as computer vision \cite{he2015deep}, speech \cite{cui2017embeddingbased}, and natural language processing \cite{vaswani2017attention}. The remarkable success of these models is driven by the huge amounts of data available and the increase in model size. As models and data grow in size, the process of training the models becomes very computation-intensive and thus time-consuming. In order to mitigate this problem, large-scale models are typically trained on a cluster of distributed workers to utilize the computing power of multiple workers. Distributed learning can be mathematically formulated as an optimization problem as follows:

\begin{align}
	\min_{\theta \in \mathbb{R}^{d}} f(\boldsymbol{\theta}) &= \frac{1}{M} \sum_{m=1}^{M} f_{m}(\boldsymbol{\theta}) \\
	f_{m}(\boldsymbol{\theta}) &= \mathop{\mathbb{E}}_{\mathbf{x}^{n} \sim \mathcal{D}^{m}} f(\boldsymbol{\theta}; \mathbf{x}^{n}) ,
\end{align}

where $\boldsymbol{\theta} \in \mathbb{R}^{n}$ denotes the model parameters to be learned, $M$ is the number of workers, $\mathcal{D}^{m}$ denotes the local data distribution at worker $m$, and $f(\boldsymbol{\theta}; \mathbf{x})$ is the loss function of the model on input $\mathbf{x}$. At each iteration, workers train a local copy of the model with a uniformly sampled subset of the total data. We assume the standard optimization setting where every worker $m$ can locally observe an unbiased stochastic gradient $\mathbf{g}^{m}_{t}$, such that $ \mathbb{E}[\mathbf{g}^{m}_{t}] = \nabla f_{m}(\boldsymbol{\theta}_{t})$ at iteration $t$. Workers sample the local stochastic gradients $\mathbf{g}^{m}_{t}$ in parallel, and they synchronize them by averaging among the workers $\mathbf{g}_{t} = \frac{1}{M} \sum_{m=1}^{M} \mathbf{g}^{m}_{t}$. The model parameters are then iteratively updated using Stochastic Gradient Descent (SGD) or one of its variants with $\boldsymbol{\theta}_{t+1} = \boldsymbol{\theta}_{t}- \eta_{t} \mathbf{g}_{t}$, where $\eta_{t}$ is the learning rate at iteration $t$.

With the increase in the number of workers, the time taken for computation significantly reduces. However, with the increase in the number of workers, the time incurred for synchronizing gradients and parameters increases. This communication time becomes a bottleneck in the training process and even can nullify the savings in computation time \cite{alistarh2018convergence}\cite{suresh2017distributed}. This bottleneck becomes detrimental when distributed learning is performed on the edge, where devices are connected via a constrained network. Many methods have been proposed to mitigate these issues, which can be broadly classified into two categories - (i) \textit{Quantization of gradients}, where workers locally quantize the gradient to a lower number of bits before communication and (ii) \textit{Sparsification of gradients}, where workers locally select a subset of gradient coordinates and communicate these values. 

Though the schemes developed on top of these methods reduce the bits communicated at every communication significantly, most of the methods fail to support parallel aggregation. Quite often, the compressed gradient cannot be directly added without decompressing it first. For example, the output of quantization in \cite{alistarh2017qsgd} is a 3-tuple, and the output of sparsification in \cite{Aji_2017} is a 2-tuple. Common gradient aggregation strategies such as coordinate-wise addition are not defined for tuples. \cite{vogels2020powersgd} identifies such gradient compression schemes whose outputs can not be hierarchically added to be non-linear. In such cases, each worker has to gather the compressed outputs of every other worker, decompress each gradient, and add them to obtain the aggregated gradient. This type of aggregation where each worker collects information from every other worker is called an all-gather operation. Modern collective communication libraries like NVIDIA NCCL provide efficient hierarchical parallel aggregation primitives such as all-reduce for dense vectors. Using all-gather limits the scalability of the algorithms as the communication times scales as $\mathcal{O}(log M)$ for all-reduce and $\mathcal{O}(M)$ for all-gather with $M$ workers. \cite{yu2018gradiveq} showed that in order to utilize parallel aggregation for compressed domain gradients, the compression function should be commutative with the gradient aggregation. Compression functions that satisfy this property require a single decompression operation at the end of all-reduce, whereas those which does not will have $M$ number of decompression operations at the end of all-gather. We can see that distributed learning with all-gather does not scale well with a large number of workers compared with all-reduce.

\paragraph{Contributions} We present a set of all-reduce compatible gradient compression algorithms -- QSGDMaxNorm Quantization, QSGDMaxNormMultiScale Quantization, and its sparsified variants. We establish upper bounds on the variance introduced by the quantization schemes and prove its convergence for smooth convex functions. We empirically evaluate the performance of the compression methods by training deep neural networks on the CIFAR10 dataset. We examine the performance of training ResNet50 (computation intensive) model and VGG16 (communication intensive) model with and without the compression methods. We also compare the scalability of these methods with the increase in the number of workers. 

\section{Related Work}
Research has proposed different solutions to tackle the problem of communication bottleneck. We focus on approaches related to gradient compression that aims to reduce the cost of communication. We look at the popular approaches of Quantization and Sparsification for gradient compression. 

\paragraph{Quantization} Typically, the gradient values are represented and communicated in 32-bit floating-point representation. Reducing the number of bits used to represent these values by quantizing them to lower-precision has been shown to reduce the cost of communication. \cite{seide2014-bit} proposed 1-bit SGD and \cite{bernstein2018signsgd} proposed SignSGD, where the gradient values are reduced to their one-bit representation. Though they show promise in reducing the communication overhead, the quantization error may impair the rate of convergence. Multi-bit unbiased quantization studied in \cite{alistarh2017qsgd} and \cite{wu2018error} allows the flexibility to trade-off between the communication cost and the convergence rate by choosing the number of bits for quantized representation. \cite{mayekar2019ratq} presented an adaptive quantization scheme where the gradients were randomly rotated and then were uniformly quantized with dynamic ranges for each coordinate. \cite{mayekar2020limits} derived lower bounds on the number of bits required to ensure the standard convergence rate of SGD and presented quantization schemes that almost attains these bounds. \cite{wu2018error} and \cite{tang2020doublesqueeze} incorporated the history of quantization error at the current quantization step. Other approaches in quantization schemes include gradient quantization to three levels \cite{wen2017terngrad}, and the quantization of gradient differences \cite{mishchenko2019distributed}.

\paragraph{Sparsification} Sparsification approaches selects and communicates a subset of the gradient coordinates. To avoid losing information, the gradients which are not communicated are accumulated over iterations and are communicated at some point. \cite{Strom2015ScalableDD} proposed sparsification along with threshold quantization to send the gradients larger than a predefined constant threshold. However, choosing a proper threshold is hard in practice since different models could exhibit different behavior. \cite{3018874} proposed to choose a fixed proportion of positive and negative gradient update and communicate them. \cite{Aji_2017} presented Gradient Dropping to sparsify the gradients by truncating the smallest gradient components based on their absolute value. Other approaches in sparsification schemes include communicating the Top K values of gradients \cite{alistarh2018convergence}, employing momentum correction \cite{lin2020deep}, and random sparsification \cite{wangni2017gradient}.

\section{Preliminaries}\label{sec:prel}
We introduce notations and definitions that are standard in convex optimization. We denote vectors with bold letters, such as $\mathbf{x}$. $x_i$ represents the $i^{th}$ coordinate of $\mathbf{x}$. We denote the sign of $t \in \mathbb{R}$ by $sign(t)$ ($-1$ if $t < 0$, or $0$ if $t = 0$, and $1$ if $t > 0$). Throughout $log$ denotes base-2 logarithm.

\begin{definition}[Convexity]
	We say that a function $f: \mathbb{R}^{n} \rightarrow \mathbb{R}$ is convex if and only if
	\begin{equation}
		f(y) \geq f(x) + \langle \nabla f(x), y - x \rangle , \forall \mathbf{x}, \mathbf{y} \in \mathbb{R}^n .
	\end{equation}
\end{definition}

\begin{definition}[L-smoothness]
	We say that a function $f: \mathbb{R}^{n} \rightarrow \mathbb{R}$ is L-smooth if its gradient is
	L-Lipschitz continuous,
	\begin{equation}
		\Vert \nabla f(\mathbf{x}) - \nabla f(\mathbf{y}) \rVert_{2} \leq L \Vert \mathbf{x} - \mathbf{y} \rVert_{2}, \forall \mathbf{x}, \mathbf{y} \in \mathbb{R}^n .
	\end{equation}
\end{definition}

We consider the standard stochastic gradient descent setup for convex functions \cite{bubeck2015convex}. Let $\Theta \subseteq \mathbb{R}^{n}$ be a closed and convex set. We wish to minimize $f: \Theta \rightarrow \mathbb{R}$, where $f$ is an unknown, differentiable, convex, and L-smooth function. We assume access to stochastic gradients $\tilde{g}$ of $f$, such that $\mathbb{E}[\tilde{g}(\boldsymbol{\theta})] = \nabla f(\boldsymbol{\theta})$. We assume the stochastic gradients have a second moment upper bound $B$, such that $\mathbb{E}[\Vert \tilde{g} (\boldsymbol{\theta}) \rVert^{2}_{2}] \leq B$ for all $\boldsymbol{\theta} \in \Theta$. We assume the stochastic gradients have a variance upper bound $\sigma^{2}$, such that $\mathbb{E}[\Vert \tilde{g} (\boldsymbol{\theta}) - \nabla f(\boldsymbol{\theta}) \rVert^{2}_{2}] \leq \sigma^{2}$ for all $\boldsymbol{\theta} \in \Theta$.

Under this setup, we have the following result:
\begin{theorem}\label{the:1} 
	\cite{bubeck2015convex} Let $\Theta \subseteq \mathbb{R}^n$ be a closed and convex set. Let  $f: \Theta \rightarrow \mathbb{R}$ be an unknown, convex, and L-smooth function. Let $\boldsymbol{\theta}_{0} \in \Theta$ be an arbitrary initial point, and let $R^{2} = sup_{\boldsymbol{\theta} \in \Theta} \Vert \boldsymbol{\theta} - \boldsymbol{\theta}_{0} \rVert^{2}$. Given repeated and independent access to stochastic gradients with variance upper bound $\sigma^{2}$, projected SGD executed for T iterations with constant step-size $\eta_{t} = \frac{1}{L + \nicefrac{1}{\gamma}}$, where $\gamma= \frac{R}{\sigma}\sqrt{\frac{2}{T}}$ satisfies,
	
	\begin{equation}
		\mathbb{E}\left[ f(\frac{1}{T} \sum_{t=0}^{T} \boldsymbol{\theta}_{t})\right]  - \min_{\boldsymbol{\theta} \in \Theta} f(\boldsymbol{\theta})
		\leq R \sqrt{\frac{2 \sigma^{2}}{T}} + \frac{LR^{2}}{T} .
	\end{equation}
\end{theorem}

For a distributed data parallel setup with $M$ workers, we have the following result:

\begin{corollary}\label{cor:1}
	\cite{alistarh2017qsgd} Let $\Theta$, f, L, $\boldsymbol{\theta}_{0}$, $\gamma$, and R be as defined in Theorem \ref{the:1} and let $\epsilon>0$. Suppose we execute projected SGD on M processors, each with access to independent stochastic gradients of $f$with second moment upper bound $B$, with step-size $\eta_{t} = \frac{1}{L + \frac{\sqrt{M}}{\gamma}}$. Then for $T= O\left( R^2 \cdot \max (\frac{2B}{K\epsilon^2}, \frac{L}{\epsilon}) \right)$ iterations,
	\begin{equation}
		\mathbb{E}\left[ f(\frac{1}{T} \sum_{t=0}^{T} \boldsymbol{\theta}_{t})\right]  - \min_{\boldsymbol{\theta} \in \Theta} f(\boldsymbol{\theta})
		\leq \epsilon .
	\end{equation}
\end{corollary}

\section{Algorithms}
Consider a distributed learning setting with $M$ workers. Each worker $m$ has access to an unbiased stochastic gradient $\mathbf{g}^{m}_{t}$, such that $\mathbb{E}[\mathbf{g}^{m}_{t}] = \nabla f_{m}(\boldsymbol{\theta}_{t})$ at iteration $t$. Note that we consider the cost of compression and decompression of gradients, if any, to be included in the communication costs.


\subsection{QSGDMaxNorm Quantization} \label{sec:qsgd_mn}
We present a stochastic uniform quantization function similar to the quantization functions in \cite{alistarh2017qsgd}. The quantization function is given by $Q_{s}(\mathbf{v}, \Vert \mathbf{w} \rVert_{2})$, where the scale $s \geq 1$ is a tunable parameter, corresponding to the number of quantization levels. Intuitively, we are defining $s$ uniformly distributed levels between 0 and 1, and are stochastically quantizing each coordinate such that the resulting quantized vector is unbiased. The value $log(s)$ represents the precision or the number of bits required to represent each coordinate when $s$ quantization levels are used.

Consider any $\mathbf{v} \in \{\mathbf{g}^{m}_{t}: m=1:M\}$ at a given iteration $t$, where each $\mathbf{g}^{m}_{t} \in \mathbb{R}^{n}$. We define $\mathbf{w}$ as the stochastic gradient with the maximum $L_2$ norm among the workers. Thus, $\Vert \mathbf{w} \rVert_{2} = \max_{m} \Vert \mathbf{g}^{m}_{t} \rVert_{2}$. For $\mathbf{v} \neq \mathbf{0}$, $Q_{s}(\mathbf{v}, \Vert \mathbf{w} \rVert_{2})$ is defined,
\vspace*{-1.25\abovedisplayskip}
\begin{equation}
	Q_{s}(v_{i}, \Vert \mathbf{w} \rVert_{2}) = \Vert \mathbf{w} \rVert_{2} \cdot sign(v_{i}) \cdot \xi_{i} (\mathbf{v}, \Vert \mathbf{w} \rVert_{2}, s) , 
\end{equation}
where $\xi_{i} (\mathbf{v}, \Vert \mathbf{w} \rVert_{2}, s)$ are independent random variables defined as below. Let $0 \leq l < s$ be an integer such that, $|v_{i}|/\Vert \mathbf{w} \rVert_{2} \in [l/s, (l+1)/s)$. Then,
\begin{equation}
	\xi_{i} (\mathbf{v}, \Vert \mathbf{w} \rVert_{2}, s)=
	\begin{cases}
		l/s, & \text{with prob $1-p(\frac{|v_{i}|}{\Vert \mathbf{w} \rVert_{2}}, s)$}\\
		(l+1)/s, & \text{otherwise}
	\end{cases} ,
\end{equation}
where $p(a, s) = as - l$ for $a \in [0,1]$. For $\mathbf{v} = \mathbf{0}$, we define $Q_{s}(\mathbf{v}) = \mathbf{0}$. After quantization, we only need $r = \lceil log(s) \rceil + 1$ bits to represent each coordinate. The scaling factor is typically represented in 32-bit floating point representation. Hence, the total communication cost per iteration is $32 + nr$ bits, which is much smaller than $32n$ bits needed for original gradient.

\begin{algorithm}[!t]
	\caption{QSGDMaxNorm Quantization}\label{alg:qsgdmn}
	\begin{algorithmic}[1]
		\State \textbf{Input}: Local parameter vector $\boldsymbol{\theta}_{t}$, local data, learning rate $\eta_{t}$, and number of workers $M$
		\Procedure{QSGDMaxNorm}{$\boldsymbol{\theta}_{t}$}  
		\ForAll{worker $m$}
		\State $\mathbf{g}^{m}_{t} \leftarrow \nabla f_{m}(\boldsymbol{\theta}_{t})$  \Comment{Stochastic gradient} 
		\State $\Vert \mathbf{w} \rVert_{2} \leftarrow \max_{m} \Vert \mathbf{g}^{m}_{t} \rVert_{2}$ \Comment{MaxAllReduce Norm} 
		\State $\boldsymbol{\zeta}^{m}_{t} \leftarrow Q_{s}(\mathbf{g}^{m}_{t}, \Vert \mathbf{w} \rVert_{2})$  \Comment{Quantize gradient}
		
		\State $\boldsymbol{\zeta}_{t} \leftarrow \frac{1}{M} \sum_{m=1}^{M} \boldsymbol{\zeta}^{m}_{t}$ \Comment{AllReduce gradients}
		\State $\hat{\mathbf{g}}_{t} \leftarrow R_{s}(\boldsymbol{\zeta}_{t}, \Vert \mathbf{w} \rVert_{2})$ \Comment{Reconstruct gradient}
		
		\State $\boldsymbol{\theta}_{t+1} \leftarrow \boldsymbol{\theta}_{t} - \eta_{t} \hat{\mathbf{g}}_{t}$ \Comment{Update parameters}
		\EndFor
		\EndProcedure
	\end{algorithmic}
\end{algorithm}

At every iteration, each worker calculates $Q_{s}(\mathbf{v}, \Vert \mathbf{w} \rVert_{2})$ and expresses the output as a tuple $(\Vert \mathbf{w} \rVert_{2}, \boldsymbol{\zeta})$, where $\boldsymbol{\zeta}$ is the vector containing $sign(v_{i}) \cdot s \cdot \xi_{i} (\mathbf{v}, \Vert \mathbf{w} \rVert_{2}, s)$. We can easily verify this gradient compression function is commutative with gradient aggregation, and hence this quantization scheme is all-reduce compatible. Once the compressed gradients are reduced, the workers reconstruct back the aggregated gradient and update their model parameters. The reconstructed vector is given by,
\begin{equation}
	\hat{\mathbf{v}} = R_{s}( \boldsymbol{\zeta}, \Vert \mathbf{w} \rVert_{2})
	=  \Vert \mathbf{w} \rVert_{2} \cdot \boldsymbol{\zeta} / s .
\end{equation}

\subsection{QSGDMaxNormMultiScale Quantization} \label{sec:qsgd_mnms}
We present a multi-scale stochastic uniform quantization function extending the quantization function presented in the previous section. We can represent values with higher precision (finer quantization intervals) using larger scales compared to using smaller scales, thus reducing the error introduced by quantization. We can intelligently choose a higher scale representation for a coordinate while ensuring that the number of bits required is equal to that of the smallest scale representation. Intuitively, this method is equivalent to combining multiple QSGDMaxNorm Quantizers with different scales by choosing the highest scale representation for each coordinate which can be accommodated with the bits required for smallest scale representation. We can easily observe that the resulting quantized vector is unbiased. The quantization function is given by $Q_{\underline{s}}(\mathbf{v}, \Vert \mathbf{w} \rVert_{2})$, where $\underline{s} = \{s_{i}, i=1:N \}$ is the set of scales with each $s_{i} \geq 1$ a tunable parameter corresponding to the number of quantization levels and $N$ denoting the number of scales. The value $log(s_{i})$ represents the precision or the number of bits required to represent each coordinate when $s_{i}$ quantization levels are used.

Consider any $\mathbf{v} \in \{\mathbf{g}^{m}_{t}: m=1:M\}$ at a given iteration t, where each $\mathbf{g}^{m}_{t} \in \mathbb{R}^{n}$. We define $\mathbf{w}$ as the stochastic gradient with the maximum $L_2$ norm among the workers. Thus, $\Vert \mathbf{w} \rVert_{2} = \max_{m} \Vert \mathbf{g}^{m}_{t} \rVert_{2}$. For $\mathbf{v} \neq \mathbf{0}$, $Q_{\underline{s}}(\mathbf{v}, \Vert \mathbf{w} \rVert_{2})$ is defined,
\vspace*{-1.25\abovedisplayskip}
\begin{equation}
	Q_{\underline{s}}(v_{i}, \Vert \mathbf{w} \rVert_{2}) = \Vert \mathbf{w} \rVert_{2} \cdot sign(v_{i}) \cdot \xi_{i} (\mathbf{v}, \Vert \mathbf{w} \rVert_{2}, s^{*}_{i}),
\end{equation}
where $s^{*}_{i}$ for each coordinate $i$ is the largest scale $s \in \underline{s}$ satisfying,
\begin{equation}
	s \leq \frac{\Vert \mathbf{w} \rVert_{2}}{|v_{i}|}  \min_{j} s_{j},
\end{equation}
and $\xi_{i} (\mathbf{v}, \Vert \mathbf{w} \rVert_{2}, s)$ are independent random variables defined as below. Let $0 \leq l < s$ be an integer such that, $|v_{i}|/\Vert \mathbf{w} \rVert_{2} \in [l/s, (l+1)/s)$. Then,
\begin{equation}
	\xi_{i} (\mathbf{v}, \Vert \mathbf{w} \rVert_{2}, s)=
	\begin{cases}
		l/s, & \text{with prob $1-p(\frac{|v_{i}|}{\Vert \mathbf{w} \rVert_{2}}, s)$}\\
		(l+1)/s, & \text{otherwise}
	\end{cases},
\end{equation}
where $p(a, s) = as - l$ for $a \in [0,1]$. For $\mathbf{v} = \mathbf{0}$, we define $Q_{s}(\mathbf{v}) = \mathbf{0}$. However, any arbitrary coordinate can be in different scales across different workers, making this scheme all-reduce incompatible. We propose a technique, which we call as \textit{scale sharing}, which enables us to choose a common scale for a particular coordinate across the workers. We define $s^{*}_{i} = \min_{m} s^{*, m}_{i}$, where $s^{*, m}_{i}$ is the scale calculated for coordinate $i$ in worker $m$. This sharing process is all-reduce compatible and has an additional overhead of communicating $\lceil log(N) \rceil$ bits per coordinate. After quantization, we only need $r = \lceil log(s) \rceil + 1 + \lceil log(N) \rceil$ bits to represent each coordinate. The scaling factor is typically represented in 32-bit floating-point representation. Hence, the total communication cost per iteration is $32 + nr$ bits, which is much smaller than $32n$ bits needed for the original gradient.

\begin{algorithm}[!t]
	\caption{QSGDMaxNormMultiScale Quantization}\label{alg:qsgdmnts}
	\begin{algorithmic}[1]
		\State \textbf{Input}: Local parameter vector $\boldsymbol{\theta}_{t}$, local data, learning rate $\eta_{t}$, and number of workers $M$
		\Procedure{QSGDMaxNormMultiScale}{$\boldsymbol{\theta}_{t}$}  
		\ForAll{worker $m$}
		\State $\mathbf{g}^{m}_{t} \leftarrow \nabla f_{m}(\boldsymbol{\theta}_{t})$  \Comment{Stochastic gradient} 
		\State $\Vert \mathbf{w} \rVert_{2} \leftarrow \max_{m} \Vert \mathbf{g}^{m}_{t} \rVert_{2}$ \Comment{MaxAllReduce Norm} 
		
		\State Compute $s^{*, m}$  \Comment{Compute coordinate scales}
		\State $s^{*} \leftarrow \min_{m=1}^{M} s^{*, m}$  \Comment{Scale sharing}
		
		\State $\boldsymbol{\zeta}^{m}_{t} \leftarrow Q_{\underline{s}}(\mathbf{g}^{m}_{t}, \Vert \mathbf{w} \rVert_{2})$  \Comment{Quantize with $s^{*}$}		
		
		\State $\boldsymbol{\zeta}_{t} \leftarrow \frac{1}{M} \sum_{m=1}^{M} \boldsymbol{\zeta}^{m}_{t}$ \Comment{AllReduce gradients}
		
		\State $\hat{\mathbf{g}}_{t} \leftarrow R_{\underline{s}}(\boldsymbol{\zeta}_{t}, \Vert \mathbf{w} \rVert_{2})$ \Comment{Reconstruct with $s^{*}$}
		
		\State $\boldsymbol{\theta}_{t+1} \leftarrow \boldsymbol{\theta}_{t} - \eta_{t} \hat{\mathbf{g}}_{t}$ \Comment{Update parameters}
		\EndFor 
		\EndProcedure
	\end{algorithmic}
\end{algorithm}

At every iteration, each worker calculates the scale vector $s^{*,m}$ consisting of scales for every coordinate and performs scale sharing. The workers then calculates $Q_{\underline{s}}(\mathbf{v}, \Vert \mathbf{w} \rVert_{2})$ and expresses the output as a tuple $(\Vert \mathbf{w} \rVert_{2}, \boldsymbol{\zeta})$, where $\boldsymbol{\zeta}$ is the vector containing $sign(v_{i}) \cdot s^{*}_{i} \cdot \xi_{i} (\mathbf{v}, \Vert \mathbf{w} \rVert_{2}, s^{*}_{i})$. We can easily verify this gradient compression function is commutative with gradient aggregation, and hence this quantization scheme is all-reduce compatible. Once the compressed gradients are reduced, the workers reconstruct back the aggregated gradient and update their model parameters. The reconstructed vector is given by,
\begin{equation}
	\hat{\mathbf{v}} = R_{\underline{s}}(\boldsymbol{\zeta}, \Vert \mathbf{w} \rVert_{2})
	=  \Vert \mathbf{w} \rVert_{2} \cdot \boldsymbol{\zeta} \cdot / s^{*}.
\end{equation}
where $\cdot /$ denotes element-wise division.

\subsection{GlobalRandKMaxNorm Compression} \label{sec:grandk_mn}
We sparsify the gradients by choosing $K$ coordinates uniformly at every iteration and apply the QSGDMaxNorm Quantization scheme on them to further reduce the communication.

\subsection{GlobalRandKMaxNormMultiScale Compression} \label{sec:grandk_mnms}
We sparsify the gradients by choosing $K$ coordinates uniformly at every iteration and apply the QSGDMaxNormMultiScale Quantization scheme on them to further reduce the communication.

\section{Convergence Guarantees}
We provide theoretical convergence guarantees to the quantization schemes we introduced in the previous section for smooth and convex functions. We assume that all assumptions listed in Section \ref{sec:prel} hold. Proofs are provided in the appendix of this paper. 

\subsection{QSGDMaxNorm Quantization}
Consider the QSGDMaxNorm Quantization presented in Section \ref{sec:qsgd_mn}. We show the following results.
\begin{lemma}[Unbiasedness and Variance bound]\label{lem:1}
	For any $\mathbf{v} \in \mathbb{R}^{n}$, we have $\mathbb{E}[Q_{s}(\mathbf{v}, \Vert \mathbf{w} \rVert_{2})] = \mathbf{v}$ and $\mathbb{E}[\Vert Q_{s}(\mathbf{v}, \Vert \mathbf{w} \rVert^{2}_{2}) - \mathbf{v} \rVert^{2}_{2}] \leq (1 + min(\frac{n}{s^{2}}, \frac{\sqrt{n}}{s})) \lVert \mathbf{w} \rVert^{2}_{2}$.
\end{lemma}

Combining the results of Lemma \ref{lem:1} with Theorem \ref{the:1} and Corollary \ref{cor:1}, we have the following guarantee for QSGDMaxNorm Quantization:
\begin{theorem}\label{the:2}
	Let $\Theta, f, L, \boldsymbol{\theta}_{0},$ and $R$ be as in Theorem \ref{the:1}. Let $\epsilon > 0$ and $\gamma = \frac{R}{\tilde{B}} \sqrt{\frac{2}{T}}$ with $\tilde{B} = \left(2 + \min \left( \frac{n}{s^{2}}, \frac{\sqrt{n}}{s} \right) \right)B$. Suppose we run QSGDMaxNorm with $s$ quantization levels on $M$ processors parallelly, each having access to independent stochastic gradients of $f$ with a second moment upper bound $B$, with step-size $\eta_{t} = \frac{1}{L + \frac{\sqrt{M}}{\gamma}}$. Then, we have for $T= O\left(R^2 \cdot \max (\frac{2 \tilde{B}}{K\epsilon^2}, \frac{L}{\epsilon}) \right)$ iterations,
	\begin{equation}
		\mathbb{E}\left[ f(\frac{1}{T} \sum_{t=0}^{T} \boldsymbol{\theta}_{t})\right]  - \min_{\boldsymbol{\theta} \in \Theta} f(\boldsymbol{\theta})
		\leq \epsilon .
	\end{equation}
\end{theorem}


\subsection{QSGDMaxNormMultiScale Quantization}
Consider the QSGDMaxNormMultiScale Quantization presented in Section \ref{sec:qsgd_mnms}. We show the following results.

\begin{lemma}[Unbiasedness and Variance bound]\label{lem:2}
	For any $\mathbf{v} \in \mathbb{R}^{n}$, we have $\mathbb{E}[Q_{\underline{s}}(\mathbf{v}, \Vert \mathbf{w} \rVert_{2})] = \mathbf{v}$ and $\mathbb{E}[\Vert Q_{\underline{s}}(\mathbf{v}, \Vert \mathbf{w} \rVert^{2}_{2}) - \mathbf{v} \rVert^{2}_{2}] \leq (1 + min(\frac{n}{\hat{s}^{2}}, \frac{\sqrt{n}}{\hat{s}})) \lVert \mathbf{w} \rVert^{2}_{2}$, where $\hat{s} = \min_{s \in \underline{s}} s$.
\end{lemma} 


Combining the results of Lemma \ref{lem:2} with Theorem \ref{the:1} and Corollary \ref{cor:1}, we have the following guarantee for QSGDMaxNormMultiScale Quantization:
\begin{theorem}\label{the:3}
	Let $\Theta, f, L, \boldsymbol{\theta}_{0},$ and $R$ be as in Theorem \ref{the:1}. Let $\epsilon > 0$ and $\gamma = \frac{R}{\tilde{B}} \sqrt{\frac{2}{T}}$ with $\tilde{B} = \left(2 + \min \left( \frac{n}{\hat{s}^{2}}, \frac{\sqrt{n}}{\hat{s}} \right) \right)B$, where $\hat{s} = \min_{s \in \underline{s}} s$. Suppose we run QSGDMaxNormMultiScale with $\underline{s}$ set of quantization levels on $M$ processors parallelly, each having access to independent stochastic gradients of $f$ with a second moment upper bound $B$, with step-size $\eta_{t} = \frac{1}{L + \frac{\sqrt{M}}{\gamma}}$. Then, we have for $T= O\left(R^2 \cdot \max (\frac{2 \tilde{B}}{K\epsilon^2}, \frac{L}{\epsilon}) \right)$ iterations,
	\begin{equation}
		\mathbb{E}\left[ f(\frac{1}{T} \sum_{t=0}^{T} \boldsymbol{\theta}_{t})\right]  - \min_{\boldsymbol{\theta} \in \Theta} f(\boldsymbol{\theta})
		\leq \epsilon .
	\end{equation}
\end{theorem}


\section{Experiments}
We implement our code in PyTorch using NVIDIA NCCL communication library. Unfortunately, PyTorch or NCCL do not support tensors of arbitrary bit length. As of writing this paper, the framework supports tensors of data types: signed and unsigned 8-bit integer and 16-bit floating-point real number. We omit data types with a bit width not smaller than the default floating-point representation of gradients (32-bit). Obviously, it does not reduce the communication cost but adds additional computational overhead. \cite{bernstein2019signsgd} implemented a bit-packing scheme to efficiently pack the sign bits into a tensor to overcome this issue. In our experiments, we observed that even though bit-packing/encoding methods provide additional savings in bits communicated, the time taken to pack and unpack is significantly large. Hence, we chose not to perform any such methods and use the closest representation available.

We validate our approach with the image classification on the CIFAR10 dataset. We examine the performance of training ResNet50 (computation intensive) model and VGG16 (communication intensive) model. We train for 150 epochs with a batch size of 128 per worker (weak scaling). We use SGD optimizer with a momentum of $0.9$ and weight decay of $5 \times 10^{-4}$ in conjunction with Cosine Annealing LR \cite{loshchilov2017sgdr} scheduler. We implement two-scale compression schemes as a specific case for the multi-scale compressors explained in Section \ref{sec:qsgd_mnms} and Section \ref{sec:grandk_mnms}. Multi-scale compressors can be easily extended from these two-scale compressor implementations. We repeat the experiments five times and plot the mean and standard deviation for all graphs.

We begin with an explanation of the notations used for the plot legends. \textit{AllReduce-SGD} corresponds to the default gradient aggregation. \textit{QSGD-MN} and \textit{GRandK-MN} corresponds to \textit{QSGDMaxNorm Quantization} (Section \ref{sec:qsgd_mn}) and \textit{GlobalRandKMaxNorm Compression} (Section \ref{sec:grandk_mn}) respectively. The precision or number of bits used for the representation follows it. \textit{QSGD-MN-TS} and \textit{GRandK-MN-TS} corresponds to \textit{QSGDMaxNormMultiScale Quantization} (Section \ref{sec:qsgd_mnms}) and \textit{GlobalRandKMaxNormMultiScale Compression} (Section \ref{sec:grandk_mnms}) respectively, with two scales of compression. The precision or number of bits used for the representation of the two scales follows it. For the sparsified schemes, we choose the value of K as $10000$ for all the experiments. We compare our methods with a recent all-reduce compatible compression scheme \textit{PowerSGD} \cite{vogels2020powersgd} for Rank-1 and Rank-2 compression in addition to the native PyTorch aggregation. 

For single-scale schemes based on Section \ref{sec:qsgd_mn} and Section \ref{sec:grandk_mn}, we compare the performance of the schemes as we vary the precision or the number of bits as \{8, 4, 2\}. For multi-scale schemes based on Section \ref{sec:qsgd_mnms} and Section \ref{sec:grandk_mnms}, we compare the performance of the schemes as we vary the precision or the number of bits as \{(8, 12), (6, 10), (4, 8), (2, 6)\}. Intuitively, one would expect the performance of the schemes to increase with precision and multiple scales.

\subsection{Benchmarking Methods}
\begin{figure}[!h]
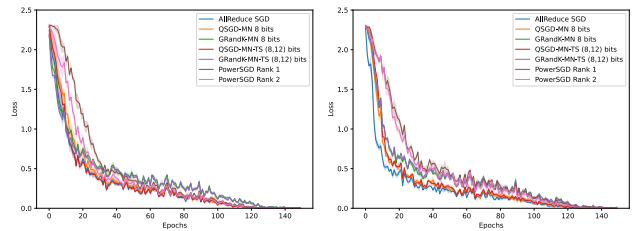

	\centering
	\begin{subfigure}{0.5\linewidth}
		\includesvg[width=\linewidth]{./imgs/CompareAll/CompareAll_loss_ResNet50.svg}
		\caption{For ResNet50}
		\label{fig:loss_epocha}
	\end{subfigure}%
	\begin{subfigure}{0.5\linewidth}
		\includesvg[width=\linewidth]{./imgs/CompareAll/CompareAll_loss_VGG16.svg}
		\caption{For VGG16}
		\label{fig:loss_epochb}
	\end{subfigure}
	\caption{Benchmarking Methods: Loss vs Epoch}
	\label{fig:loss_epoch}
\end{figure}

\begin{figure}[!h]
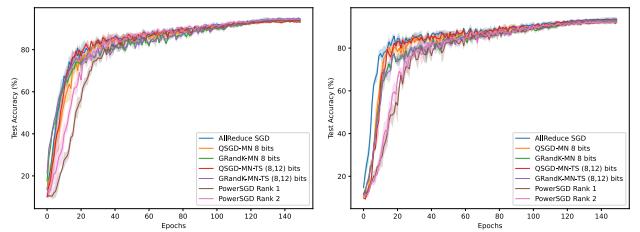

	\centering
	\begin{subfigure}{0.5\linewidth}
		\includesvg[width=\linewidth]{./imgs/CompareAll/CompareAll_top1_ResNet50.svg}
		\caption{For ResNet50}
		\label{fig:top1_epocha}
	\end{subfigure}%
	\begin{subfigure}{0.5\linewidth}
		\includesvg[width=\linewidth]{./imgs/CompareAll/CompareAll_top1_VGG16.svg}
		\caption{For VGG16}
		\label{fig:top1_epochb}
	\end{subfigure}
	\caption{Benchmarking Methods: Accuracy vs Epoch}
	\label{fig:top1_epoch}
\end{figure}

We compare the performance of our schemes with the native PyTorch aggregation and PowerSGD. Figures \ref{fig:loss_epoch} and \ref{fig:top1_epoch} show the training loss and test accuracy respectively, over the epochs during training. From figures \ref{fig:loss_epocha} and \ref{fig:top1_epocha}, we can observe that our compression schemes perform as good as the AllReduce-SGD for ResNet50 architecture. We can observe that initially the sparsified methods perform better -- even than AllReduce-SGD -- but the non-sparsified methods perform better with time. From figure \ref{fig:loss_epochb} and \ref{fig:top1_epochb}, we can observe that even though initially our compression schemes perform worse compared to Allreduce-SGD, with time, they perform as good as the Allreduce-SGD for VGG16 architecture. 
We can observe that all our methods significantly outperform PowerSGD for both ResNet50 and VGG16 architectures. This can be explained by the fact that PowerSGD uses a one-step power iteration method, which can introduce a large compression error. We can also note that the two-scale QSGD-MN-TS performs better than the single-scale QSGD-MN -- due to its reduced quantization error -- especially once the training has passed its transient state.

\subsection{QSGDMaxNorm Quantization}
\begin{figure}[!t]
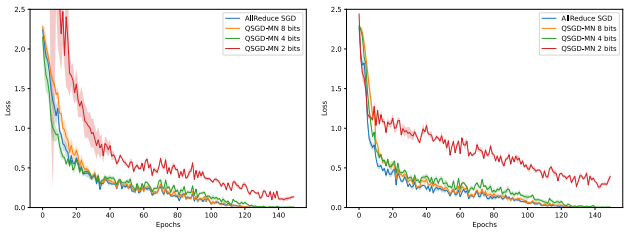

	\centering
	\begin{subfigure}{0.5\linewidth}
		\includesvg[width=\linewidth]{./imgs/QSGDMN/QSGDMN_loss_ResNet50.svg}
		\caption{For ResNet50}
		\label{fig:loss_epoch_qsgdmna}
	\end{subfigure}%
	\begin{subfigure}{0.5\linewidth}
		\includesvg[width=\linewidth]{./imgs/QSGDMN/QSGDMN_loss_VGG16.svg}
		\caption{For VGG16}
		\label{fig:loss_epoch_qsgdmnb}
	\end{subfigure}
	\caption{QSGDMaxNorm: Loss vs Epoch}
	\label{fig:loss_epoch_qsgdmn}
\end{figure}

\begin{figure}[!t]
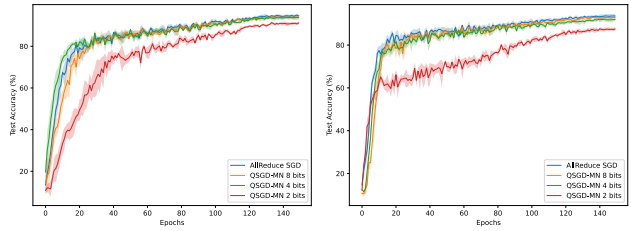

	\centering
	\begin{subfigure}{0.5\linewidth}
		\includesvg[width=\linewidth]{./imgs/QSGDMN/QSGDMN_top1_ResNet50.svg}
		\caption{For ResNet50}
		\label{fig:top1_epoch_qsgdmna}
	\end{subfigure}%
	\begin{subfigure}{0.5\linewidth}
		\includesvg[width=\linewidth]{./imgs/QSGDMN/QSGDMN_top1_VGG16.svg}
		\caption{For VGG16}
		\label{fig:top1_epoch_qsgdmnb}
	\end{subfigure}
	\caption{QSGDMaxNorm: Accuracy vs Epoch}
	\label{fig:top1_epoch_qsgdmn}
\end{figure}

Figures \ref{fig:loss_epoch_qsgdmn} and \ref{fig:top1_epoch_qsgdmn} show the training loss and test accuracy respectively, over the epochs during training. We can observe that QSGD-MN schemes with 8-bit and 4-bit quantization perform as good as the AllReduce-SGD for both ResNet50 and VGG16 architectures. However, QSGD-MN with 2-bit quantization quantizes the gradients too aggressively. As a result, the loss during the initial phase of training is significantly larger than the rest of the methods for ResNet50 architecture. The initial training loss for VGG16 architecture is comparable with the rest of the methods, unlike the case with ResNet50 architecture. Though the loss decreases with increasing number of epochs, there is a pronounced gap between the final loss of QSGD-MN with 2-bit quantization and the rest of the methods. This gap is more for VGG16 architecture compared to ResNet50 architecture. This effect is also reflected in the accuracy plots.

\subsection{QSGDMaxNormMultiScale Quantization}
\begin{figure}[!b]
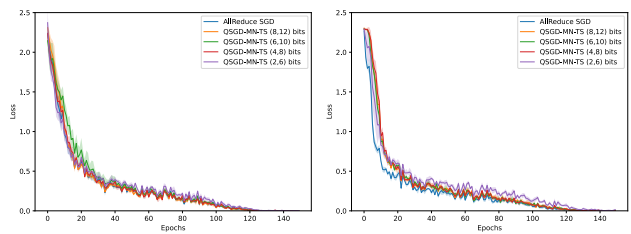

	\centering
	\begin{subfigure}{0.5\linewidth}
		\includesvg[width=\linewidth]{./imgs/QSGDMN-TS/QSGDMN-TS_loss_ResNet50.svg}
		\caption{For ResNet50}
		\label{fig:loss_epoch_qsgdmn_tsa}
	\end{subfigure}%
	\begin{subfigure}{0.5\linewidth}
		\includesvg[width=\linewidth]{./imgs/QSGDMN-TS/QSGDMN-TS_loss_VGG16.svg}
		\caption{For VGG16}
		\label{fig:loss_epoch_qsgdmn_tsb}
	\end{subfigure}
	\caption{QSGDMaxNormMultiScale: Loss vs Epoch}
	\label{fig:loss_epoch_qsgdmn_ts}
\end{figure}

\begin{figure}[!t]
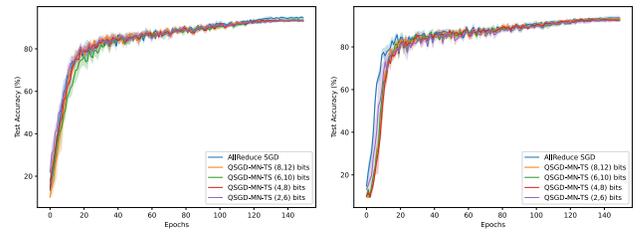

	\centering
	\begin{subfigure}{0.5\linewidth}
		\includesvg[width=\linewidth]{./imgs/QSGDMN-TS/QSGDMN-TS_top1_ResNet50.svg}
		\caption{For ResNet50}
		\label{fig:top1_epoch_qsgdmn_tsa}
	\end{subfigure}%
	\begin{subfigure}{0.5\linewidth}
		\includesvg[width=\linewidth]{./imgs/QSGDMN-TS/QSGDMN-TS_top1_VGG16.svg}
		\caption{For VGG16}
		\label{fig:top1_epoch_qsgdmn_tsb}
	\end{subfigure}
	\caption{QSGDMaxNormMultiScale: Accuracy vs Epoch}
	\label{fig:top1_epoch_qsgdmn_ts}
\end{figure}

Figures \ref{fig:loss_epoch_qsgdmn_ts} and \ref{fig:top1_epoch_qsgdmn_ts} show the training loss and test accuracy respectively, over the epochs during training. From figures \ref{fig:loss_epoch_qsgdmn_tsa} and \ref{fig:top1_epoch_qsgdmn_tsa}, we can observe that QSGD-MN-TS schemes perform as good as the AllReduce-SGD except for (6, 10) quantization for the ResNet50 architecture. From figures \ref{fig:loss_epoch_qsgdmn_tsb} and \ref{fig:top1_epoch_qsgdmn_tsb}, we can observe that QSGD-MN-TS schemes initially perform worse compared to AllReduce-SGD for the VGG16 architecture. Interestingly, the 2-bit quantization, which performed poorly in the single-scale scheme, performed in par with Allreduce-SGD in the two-scale scheme.

\subsection{GlobalRandKMaxNorm Compression}
\begin{figure}[!h]
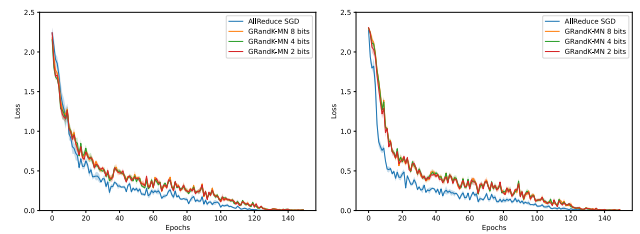

	\centering
	\begin{subfigure}{0.5\linewidth}
		\includesvg[width=\linewidth]{./imgs/GRandKMN/GRandKMN_loss_ResNet50.svg}
		\caption{For ResNet50}
		\label{fig:loss_epoch_grandka}
	\end{subfigure}%
	\begin{subfigure}{0.5\linewidth}
		\includesvg[width=\linewidth]{./imgs/GRandKMN/GRandKMN_loss_VGG16.svg}
		\caption{For VGG16}
		\label{fig:loss_epoch_grandkb}
	\end{subfigure}
	\caption{GlobalRandKMaxNorm: Loss vs Epoch}
	\label{fig:loss_epoch_grandk}
\end{figure}

\begin{figure}[!h]
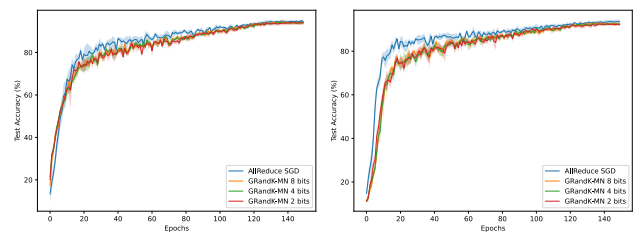

	\centering
	\begin{subfigure}{0.5\linewidth}
		\includesvg[width=\linewidth]{./imgs/GRandKMN/GRandKMN_top1_ResNet50.svg}
		\caption{For ResNet50}
		\label{fig:top1_epoch_grandka}
	\end{subfigure}%
	\begin{subfigure}{0.5\linewidth}
		\includesvg[width=\linewidth]{./imgs/GRandKMN/GRandKMN_top1_VGG16.svg}
		\caption{For VGG16}
		\label{fig:top1_epoch_grandkb}
	\end{subfigure}
	\caption{GlobalRandKMaxNorm: Accuracy vs Epoch}
	\label{fig:top1_epoch_grandk}
\end{figure}

Figures \ref{fig:loss_epoch_grandk} and \ref{fig:top1_epoch_grandk} show the training loss and test accuracy respectively, over the epochs during training. From figures \ref{fig:loss_epoch_grandka} and \ref{fig:top1_epoch_grandka}, we can observe that though GRandK-MN schemes perform better than AllReduce-SGD initially, with time, the performance becomes worse for the ResNet50 architecture. From figures \ref{fig:loss_epoch_grandkb} and \ref{fig:top1_epoch_grandkb}, we can observe that GRandK-MN schemes perform worse compared to AllReduce-SGD throughout the training for the VGG16 architecture. However, the performance improves towards the end of training. The performance is fairly resilient to the precision used since we communicate a very small subset of the gradient.

\subsection{GlobalRandKMaxNormMultiScale Compression}
\begin{figure}[!h]
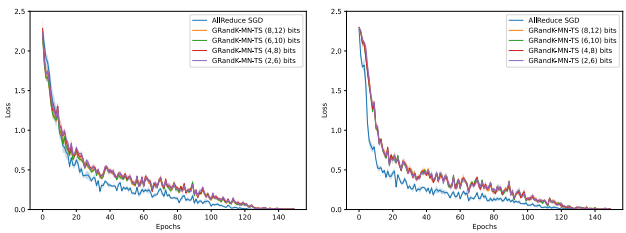

	\centering
	\begin{subfigure}{0.5\linewidth}
		\includesvg[width=\linewidth]{./imgs/GRandKMN-TS/GRandKMN-TS_loss_ResNet50.svg}
		\caption{For ResNet50}
		\label{fig:loss_epoch_grandk_tsa}
	\end{subfigure}%
	\begin{subfigure}{0.5\linewidth}
		\includesvg[width=\linewidth]{./imgs/GRandKMN-TS/GRandKMN-TS_loss_VGG16.svg}
		\caption{For VGG16}
		\label{fig:loss_epoch_grandk_tsb}
	\end{subfigure}
	\caption{GlobalRandKMaxNormMS: Loss vs Epoch}
	\label{fig:loss_epoch_grandk_ts}
\end{figure}

\begin{figure}[!h]
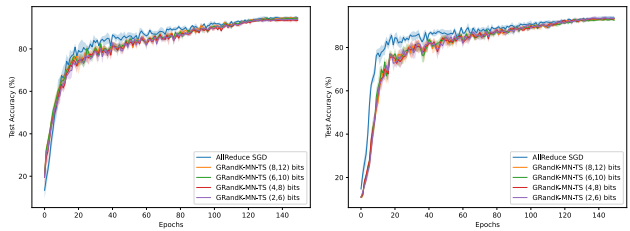

	\centering
	\begin{subfigure}{0.5\linewidth}
		\includesvg[width=\linewidth]{./imgs/GRandKMN-TS/GRandKMN-TS_top1_ResNet50.svg}
		\caption{For ResNet50}
		\label{fig:top1_epoch_grandk_tsa}
	\end{subfigure}%
	\begin{subfigure}{0.5\linewidth}
		\includesvg[width=\linewidth]{./imgs/GRandKMN-TS/GRandKMN-TS_top1_VGG16.svg}
		\caption{For VGG16}
		\label{fig:top1_epoch_grandk_tsb}
	\end{subfigure}
	\caption{GlobalRandKMaxNormMS: Accuracy vs Epoch}
	\label{fig:top1_epoch_grandk_ts}
\end{figure}

Figures \ref{fig:loss_epoch_grandk_ts} and \ref{fig:top1_epoch_grandk_ts} show the training loss and test accuracy respectively, over the epochs during training. We can draw similar observations made in the previous section.

\subsection{Performance Modeling}
We use the performance model developed in \cite{wen2017terngrad} to analyze the scalability of the compression schemes. We perform lightweight profiling on an AWS p3.8xlarge instance with 4 NVIDIA V100 GPUs. This instance supports NVLink GPU Peer to Peer Communication and has a network bandwidth of 10 Gbps. We then use this analytical model to study the throughput (images processed per second) for a cluster of 32 nodes, each equipped with 4 NVIDIA V100 GPUs with NVLink GPU interconnect. 

\begin{figure}[!t]
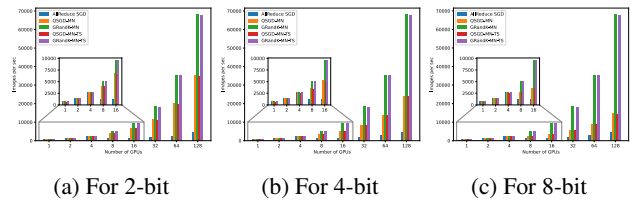

	\centering
	\begin{subfigure}{0.33\linewidth}
		\includesvg[width=\linewidth]{./imgs/Performance_Modelling/2/1GBPs/2_1_performance_modelling_ResNet50_P3.svg}
		\caption{For 2-bit}
		\label{fig:perfmodel_resnet_1gbpsa}
	\end{subfigure}%
	\begin{subfigure}{0.33\linewidth}
		\includesvg[width=\linewidth]{./imgs/Performance_Modelling/4/1GBPs/4_1_performance_modelling_ResNet50_P3.svg}
		\caption{For 4-bit}
		\label{fig:perfmodel_resnet_1gbpsb}
	\end{subfigure}%
	\begin{subfigure}{0.33\linewidth}
		\includesvg[width=\linewidth]{./imgs/Performance_Modelling/8/1GBPs/8_1_performance_modelling_ResNet50_P3.svg}
		\caption{For 8-bit}
		\label{fig:perfmodel_resnet_1gbpsc}
	\end{subfigure}
	\caption{Performance: ResNet50 with 1 Gbps Ethernet}
	\label{fig:perfmodel_resnet_1gbps}
\end{figure}

\begin{figure}[!t]
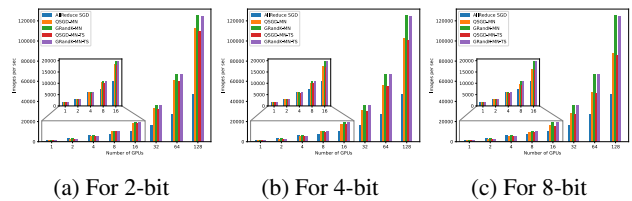

	\centering
	\begin{subfigure}{0.33\linewidth}
		\includesvg[width=\linewidth]{./imgs/Performance_Modelling/2/10GBPs/2_10_performance_modelling_VGG16_P3.svg}
		\caption{For 2-bit}
		\label{fig:perfmodel_vgg_10gbpsa}
	\end{subfigure}%
	\begin{subfigure}{0.33\linewidth}
		\includesvg[width=\linewidth]{./imgs/Performance_Modelling/4/10GBPs/4_10_performance_modelling_VGG16_P3.svg}
		\caption{For 4-bit}
		\label{fig:perfmodel_vgg_10gbpsb}
	\end{subfigure}%
	\begin{subfigure}{0.33\linewidth}
		\includesvg[width=\linewidth]{./imgs/Performance_Modelling/8/10GBPs/8_10_performance_modelling_VGG16_P3.svg}
		\caption{For 8-bit}
		\label{fig:perfmodel_vgg_10gbpsc}
	\end{subfigure}
	\caption{Performance: VGG16 with 10 Gbps Ethernet}
	\label{fig:perfmodel_vgg_10gbps}
\end{figure}

We evaluate the performance under 1 Gbps and 10 Gbps Ethernet connection between the nodes. Under each condition, we vary the number of bits of quantization in \{2, 4, 8\}. For the two-scale compressors, this corresponds to the minimum value of the scales. Figures \ref{fig:perfmodel_resnet_1gbps} and \ref{fig:perfmodel_resnet_10gbps} shows the system throughput under 1 Gbps and 10 Gbps Ethernet connections for the ResNet50 architecture. Figures \ref{fig:perfmodel_vgg_1gbps} and \ref{fig:perfmodel_vgg_10gbps} shows the system throughput under 1 Gbps and 10 Gbps Ethernet connections for VGG16 architecture. We observe that our compression schemes have substantial speedup compared to the native synchronous SGD. The throughput decreases with an increase in the number of bits used for quantization. Under low bandwidth 1 Gbps Ethernet connection, sparsified methods significantly outperform the non-sparsified methods because of reduced communication. The speedup gain in VGG16 architecture is more than of ResNet50 architecture because the former is a communication-intensive model and the latter is a computation-intensive model. Compression schemes help to increase the system throughput when communication-to-computation ratio of the model is higher and the network bandwidth is lower.

\begin{figure}[!t]
	\centering
	\begin{subfigure}{0.33\linewidth}
		\includesvg[width=\linewidth]{./imgs/Performance_Modelling/2/10GBPs/2_10_performance_modelling_ResNet50_P3.svg}
		\caption{For 2-bit}
		\label{fig:perfmodel_resnet_10gbpsa}
	\end{subfigure}%
	\begin{subfigure}{0.33\linewidth}
		\includesvg[width=\linewidth]{./imgs/Performance_Modelling/4/10GBPs/4_10_performance_modelling_ResNet50_P3.svg}
		\caption{For 4-bit}
		\label{fig:perfmodel_resnet_10gbpsb}
	\end{subfigure}%
	\begin{subfigure}{0.33\linewidth}
		\includesvg[width=\linewidth]{./imgs/Performance_Modelling/8/10GBPs/8_10_performance_modelling_ResNet50_P3.svg}
		\caption{For 8-bit}
		\label{fig:perfmodel_resnet_10gbpsc}
	\end{subfigure}
	\caption{Performance: ResNet50 with 10 Gbps Ethernet}
	\label{fig:perfmodel_resnet_10gbps}
\end{figure}

\begin{figure}[!t]
	\centering
	\begin{subfigure}{0.33\linewidth}
		\includesvg[width=\linewidth]{./imgs/Performance_Modelling/2/1GBPs/2_1_performance_modelling_VGG16_P3.svg}
		\caption{For 2-bit}
		\label{fig:perfmodel_vgg_1gbpsa}
	\end{subfigure}%
	\begin{subfigure}{0.33\linewidth}
		\includesvg[width=\linewidth]{./imgs/Performance_Modelling/4/1GBPs/4_1_performance_modelling_VGG16_P3.svg}
		\caption{For 4-bit}
		\label{fig:perfmodel_vgg_1gbpsb}
	\end{subfigure}%
	\begin{subfigure}{0.33\linewidth}
		\includesvg[width=\linewidth]{./imgs/Performance_Modelling/8/1GBPs/8_1_performance_modelling_VGG16_P3.svg}
		\caption{For 8-bit}
		\label{fig:perfmodel_vgg_1gbpsc}
	\end{subfigure}
	\caption{Performance: VGG16 with 1 Gbps Ethernet}
	\label{fig:perfmodel_vgg_1gbps}
\end{figure}

\section{Concluding Remarks}
With the increasing compute power of hardware systems, communication will soon be the bottleneck of large-scale learning. In this paper, we presented a set of all-reduce compatible gradient compression schemes which reduce the communication costs in distributed data-parallel machine learning. We establish upper bounds on the variance introduced by the proposed quantization schemes. We further establish the convergence of the quantization schemes for smooth and convex objective functions. The schemes allow to trade-off between the convergence and the communication costs by varying the number of quantization levels. We also conduct extensive experiments on practical deep learning models and evaluate the performance of the quantization schemes.

{
\appendix
\section{Discussions}
We present some generic observations and discussions on the usage of gradient compression for distributed learning.

\paragraph{Necessity of gradient compression} While gradient compression methods significantly reduce communication costs in terms of bits, the gain in terms of time may not be significant. Many gradient compression schemes such as Top-K take a significant amount of time to compress the gradients. The savings in bits may be attractive in federated learning, but they are not in data center training. Modern multi-node data center environments have upwards of 300 GBps GPU interconnect (such as NVIDIA NVLink3) and upwards of 10 Gbps Ethernet connection. Under these conditions, if the time taken for compressing and communicating the gradients is more than communicating the gradients without the compression, there is no necessity of using compression schemes.

\paragraph{All-reduce vs All-gather} The all-reduce communication primitive provides an efficient aggregation method to communicate between the workers compared to all-gather communication primitive. As a result, gradient compression schemes that are all-reduce compatible provide much better scaling. However, most of the current gradient schemes are not all-reduce compatible, which results in poor scaling with the number of workers.

\paragraph{Computation-intensive training} The computation intensive training limits the acceleration and performance gains with gradient compression. A training can be computation-intensive if we train a complex model or have a large batch size. In either case, the amount of time spent in computing can be much larger than the time spent in communication. This results in diminished performance gains when using gradient compression. On the other hand, communication-intensive training can gain significantly using gradient compression schemes. Given the rate at which hardware compute power is growing, most training in the future will be bottlenecked by communication, and gradient compression methods can play a crucial role in alleviating it.

\paragraph{Limitations of the framework} Currently, the deep learning frameworks and communication libraries support only a limited range of tensor datatypes. Any compression schemes which use a representation that is not supported will have to use the next largest representation supported. This typically results in wastage of bits and hampers the performance gain. For example, the smallest representation supported by these frameworks are 8-bit tensors. If we wish to use 2-bit or 4-bit tensors, we will have to use 8-bit tensors. One way to mitigate this issue is to pack multiple coordinates to a dense tensor -- which usually takes time and makes the scheme all-reduce incompatible. However, this issue will be resolved if the framework supports tensors of arbitrary bit-width.

\section{Time Breakdown}
We provide justifications for the substantial speedup in the distributed learning when compressors are employed. Figures \ref{fig:time_breakdowna} and \ref{fig:time_breakdownb} shows the time breakdown for various sub-processes in the training for ResNet50 architecture and VGG16 architecture respectively. We measure the times from a cluster of 4 machines, each with an NVIDIA V100 GPU. 

\begin{figure}[!t]
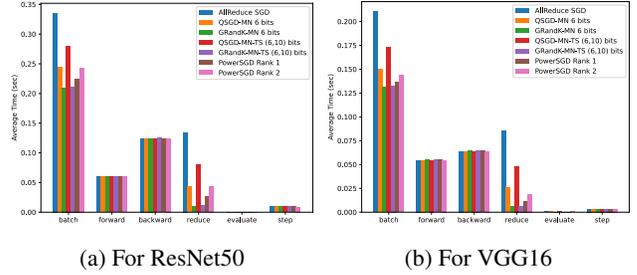

	\centering
	\begin{subfigure}{0.5\linewidth}
		\includesvg[width=\linewidth]{./imgs/Time_Breakdown/time_breakdown_ResNet50.svg}
		\caption{For ResNet50}
		\label{fig:time_breakdowna}
	\end{subfigure}%
	\begin{subfigure}{0.5\linewidth}
		\includesvg[width=\linewidth]{./imgs/Time_Breakdown/time_breakdown_VGG16.svg}
		\caption{For VGG16}
		\label{fig:time_breakdownb}
	\end{subfigure}
	\caption{Time Breakdown of Compression methods}
	\label{fig:time_breakdown}
\end{figure}

We can observe that the differences in training times arise due to the differences in the communication time. Observe that the two-scale methods have larger communication times because of two all-reduce operations of 8-bit tensors. The framework currently does not support tensors less than 8-bit as of writing this paper, and this increases the communication time for the two-scale methods. We expect that this time should decrease when the framework supports tensors of lower bit-width. We can also observe from the figures that the QSGDMaxNorm Quantization takes the same amount of time of PowerSGD compression with Rank-2 approximation, while ensuring better convergence and performance for ResNet50 architecture. QSGDMaxNorm Quantization takes slightly more time than PowerSGD compression with Rank-2 approximation for VGG16 architecture. This behavior can be explained by the fact that ResNet50 architecture has $23520842$ parameters and VGG16 architecture has $14728266$ parameters. A larger number of parameters would require more time to calculate low-rank approximations.

\section{Proof for Lemma \ref{lem:1}}
\begin{proof}
	We first show unbiasedness of the quantization. For any $\mathbf{v} \in \mathbb{R}^{n}$ and any coordinate index $i$ we have,
	\begin{align*}
		&\mathbb{E}[Q_{s}(v_{i}, \Vert \mathbf{w} \rVert_{2})] \\
		&= \mathbb{E}[\Vert \mathbf{w} \rVert_{2} \cdot sign(v_{i}) \cdot \xi_{i} (\mathbf{v}, \Vert \mathbf{w} \rVert_{2})] \\
		&= \Vert \mathbf{w} \rVert_{2} \cdot sign(v_{i}) \cdot \\	
		& \hspace{2.5em} \left( \frac{l}{s} * (1-p\left(\frac{|v_{i}|}{\Vert \mathbf{w} \rVert_{2}}, s\right)) + \frac{l+1}{s} * p\left(\frac{|v_{i}|}{\Vert \mathbf{w} \rVert_{2}}, s \right) \right) \\
		&= \Vert \mathbf{w} \rVert_{2} \cdot sign(v_{i}) \cdot \frac{|v_{i}|}{\Vert \mathbf{w} \rVert_{2}} \\
		& = v_{i} . 
	\end{align*}
	We will now show the variance upper bound. Recall the definition of $\xi_{i} (\mathbf{v}, \Vert \mathbf{w} \rVert_{2}, s)$,
	\begin{align*}		
			\xi_{i} (\mathbf{v}, \Vert \mathbf{w} \rVert_{2}, s)=
		\begin{cases}
			l/s, & \text{with prob $1-p(\frac{|v_{i}|}{\Vert \mathbf{w} \rVert_{2}}, s)$}\\
			(l+1)/s, & \text{otherwise}
		\end{cases}.
	\end{align*}
	We can show,
	\begin{align*}
		&\mathbb{E}[\xi_{i} (\mathbf{v}, \Vert \mathbf{w} \rVert_{2}, s)^{2}] \\
		&= \mathbb{E}[\xi_{i} (\mathbf{v}, \Vert \mathbf{w} \rVert_{2}, s)]^{2} + Var\left[(\xi_{i} (\mathbf{v}, \Vert \mathbf{w} \rVert_{2}, s) \right] \\ \\
		&= \frac{v^{2}_{i}}{\lVert \mathbf{w} \rVert^{2}_{2}} + \frac{1}{s^{2}} p\left(\frac{|v_{i}|}{\Vert \mathbf{w} \rVert_{2}}, s \right)\left( 1 - p\left(\frac{|v_{i}|}{\Vert \mathbf{w} \rVert_{2}}, s \right) \right) \\
		&\leq \frac{v^{2}_{i}}{\lVert \mathbf{w} \rVert^{2}_{2}} + \frac{1}{s^{2}} p\left(\frac{|v_{i}|}{\Vert \mathbf{w} \rVert_{2}}, s \right) .
	\end{align*}
	We now upper bound the $L_{2}$ norm of the quantized vector as,
	\begin{align*}		
		&\mathbb{E}[\Vert Q_{s}(\mathbf{v}, \Vert \mathbf{w} \rVert_{2}) \rVert^{2}_{2}] \\
		&=  \sum_{i=1}^{n} \mathbb{E}\left[\lVert \mathbf{w} \rVert^{2}_{2} \xi_{i} (\mathbf{v}, \Vert \mathbf{w} \rVert_{2}, s)^{2} \right] \\
		&\leq \lVert \mathbf{w} \rVert^{2}_{2} \sum_{i=1}^{n} \left[ \frac{v^{2}_{i}}{\lVert \mathbf{w} \rVert^{2}_{2}} + \frac{1}{s^{2}} p\left(\frac{|v_{i}|}{\Vert \mathbf{w} \rVert_{2}}, s \right) \right] \\
		&\leq \left(1 + \frac{1}{s^{2}} \sum_{i=1}^{n} p\left(\frac{|v_{i}|}{\Vert \mathbf{w} \rVert_{2}}, s \right) \right) \lVert \mathbf{w} \rVert^{2}_{2} \\
		&\leq \left(1 + \min \left( \frac{n}{s^{2}}, \frac{\lVert \mathbf{v} \rVert_{1}}{s \lVert \mathbf{w} \rVert_{2}}\right)\right) \lVert \mathbf{w} \rVert^{2}_{2} \\
		&\leq \left(1 + \min \left( \frac{n}{s^{2}}, \frac{\lVert \mathbf{v} \rVert_{1}}{s \lVert \mathbf{v} \rVert_{2}}\right)\right) \lVert \mathbf{w} \rVert^{2}_{2} \\
		&\leq \left(1 + \min \left( \frac{n}{s^{2}}, \frac{\sqrt{n}}{s} \right) \right)\lVert \mathbf{w} \rVert^{2}_{2} .
	\end{align*}
	Using the above results we can derive the variance upper bound for the quantized vector,
	\begin{align*}
		&\mathbb{E}[\Vert Q_{s}(\mathbf{v}, \Vert \mathbf{w} \rVert_{2}) - \mathbf{v} \rVert^{2}_{2}] \\
		&\leq  \mathbb{E}[\Vert Q_{s}(\mathbf{v}, \Vert \mathbf{w} \rVert_{2}) \rVert^{2}_{2}] - \lVert \mathbf{v} \rVert^{2}_{2} \\
		&\leq \min \left( \frac{n}{s^{2}}, \frac{\sqrt{n}}{s} \right) \lVert \mathbf{w} \rVert^{2}_{2} + \lVert \mathbf{w} \rVert^{2}_{2} - \lVert \mathbf{v} \rVert^{2}_{2} \\
		&\leq \left(1 + \min \left( \frac{n}{s^{2}}, \frac{\sqrt{n}}{s} \right) \right)\lVert \mathbf{w} \rVert^{2}_{2} .
	\end{align*}
	This result means that if $\tilde{g}(\boldsymbol{\theta})$ is a stochastic gradient with second moment upper bound $B$, then $Q_{s}(\tilde{g}(\boldsymbol{\theta}), \Vert \mathbf{w} \rVert_{2})$ is a stochastic gradient with upper bound $\alpha B$, where $\alpha = 1 + \min \left( \frac{n}{s^{2}}, \frac{\sqrt{n}}{s} \right)$.
\end{proof}

\section{Proof for Theorem \ref{the:2}}
\begin{proof}
	Let $\tilde{g}(\boldsymbol{\theta})$ and $\tilde{h}(\boldsymbol{\theta})$ be the stochastic gradient and quantized stochastic gradient respectively with  $\tilde{h}(\boldsymbol{\theta}) = Q_{s}(\tilde{g}(\boldsymbol{\theta}), \Vert \mathbf{w} \rVert_{2})$. Then we have,
	\begin{align*}
		&\mathbb{E}[\lVert \tilde{h}(\boldsymbol{\theta}) - \nabla f(\boldsymbol{\theta}) \rVert^{2}_{2}] \\
		&\leq \mathbb{E}[\lVert \tilde{h}(\boldsymbol{\theta}) - \tilde{g}(\boldsymbol{\theta}) \rVert^{2}_{2}] + \mathbb{E}[\lVert \tilde{g}(\boldsymbol{\theta}) - \nabla f(\boldsymbol{\theta}) \rVert^{2}_{2}] \\
		&\leq \left(1 + \min \left( \frac{n}{s^{2}}, \frac{\sqrt{n}}{s} \right)\right)B + B \\
		&= \left(2 + \min \left( \frac{n}{s^{2}}, \frac{\sqrt{n}}{s} \right)\right)B ,
	\end{align*}
	where we use Lemma \ref{lem:1} and the second moment upper bound assumption  of stochastic gradients.
\end{proof}

\section{Proof for Lemma \ref{lem:2}}
\begin{proof}
	The proof follows the same way as of proof for QSGDMaxNorm Quantization scheme and noting that $\hat{s} = \min_{s \in \underline{s}} s$. This result means that if $\tilde{g}(\boldsymbol{\theta})$ is a stochastic gradient with second moment upper bound $B$, then $Q_{\underline{s}}(\tilde{g}(\boldsymbol{\theta}), \Vert \mathbf{w} \rVert_{2})$ is a stochastic gradient with upper bound $\beta B$, where $\beta = 1 + \min \left( \frac{n}{\hat{s}^{2}}, \frac{\sqrt{n}}{\hat{s}} \right)$ and $\hat{s} = \min_{s \in \underline{s}} s$.
\end{proof}

\section{Proof for Theorem \ref{the:3}}
\begin{proof}
	The proof follows the same way as proof for the QSGDMaxNorm Quantization scheme.
\end{proof}

}

\section*{Acknowledgments}
The author would like to thank Himanshu Tyagi for suggesting the problem, useful discussions, and guidance throughout this work. This work has been supported by a research grant from the Robert Bosch Center for Cyber-Physical Systems at the Indian Institute of Science, Bangalore.

{
\bibliography{aaai22}
}

\end{document}